\documentclass[10pt,twocolumn,letterpaper]{article}

\usepackage{cvpr}
\usepackage{times}
\usepackage{epsfig}
\usepackage{graphicx}
\usepackage{amsmath}
\usepackage{amssymb}

\usepackage{multirow}
\usepackage{verbatim}
\usepackage{setspace}
\usepackage{array}
\usepackage{booktabs}
\usepackage{rotating}

\newcommand{\shortcite}[1]{\cite{#1}}

\newcommand{\figref}[1]{Fig.~\ref{fig:#1}}
\newcommand{\tabref}[1]{Table~\ref{tab:#1}}
\newcommand{\equref}[1]{Eqn.~\ref{equ:#1}}
\newcommand{\secref}[1]{Sec.~\ref{sec:#1}}
\newcommand{\ignorethis}[1]{}

\cvprfinalcopy 

\def\cvprPaperID{****} 

\ifcvprfinal\fi
\begin{document}

\title{Appearance Harmonization for Single Image Shadow Removal }

\author{Li-Qian Ma\\
TNList, Tsinghua University\\
Beijing, China\\
{\tt\small liqianma@qq.com}
\and
Jue Wang\\
Adobe Systems Inc.\\
\\
{\tt\small juewang@adobe.com}
\and
Eli Shechtman\\
Adobe Systems Inc.\\
\\
{\tt\small elishe@adobe.com}
\and
Kalyan Sunkavalli\\
Adobe Systems Inc.\\
\\
{\tt\small sunkaval@adobe.com}
\and
Shi-Min Hu\\
TNList, Tsinghua University\\
Beijing, China\\
{\tt\small shiminhu@tsinghua.edu.cn}
}

\maketitle

\begin{abstract}

Shadow removal is a challenging problem and previous approaches often produce de-shadowed regions that are visually inconsistent with the rest of the image. We propose an automatic {\em shadow region harmonization} approach that makes the appearance of a de-shadowed region (produced using any previous technique) compatible with the rest of the image. We use a shadow-guided patch-based image synthesis approach that reconstructs the shadow region using patches sampled from non-shadowed regions. This result is then refined based on the reconstruction confidence to handle unique textures. Qualitative comparisons over a wide range of images, and a quantitative evaluation on a benchmark dataset show that our technique significantly improves upon the state-of-the-art.

\end{abstract}

\section{Introduction}

\begin{figure*}[t]
    \centering
\begin{tabular}{c@{\hspace{0.02in}}c@{\hspace{0.02in}}c@{\hspace{0.2in}}c@{\hspace{0.02in}}c@{\hspace{0.02in}}c}
\includegraphics[height = 0.13\linewidth]{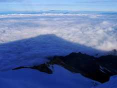} &
\includegraphics[height = 0.13\linewidth]{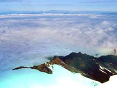} &
\includegraphics[height = 0.13\linewidth]{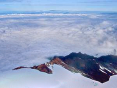} &
\includegraphics[height = 0.13\linewidth]{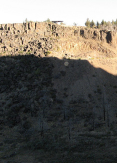} &
\includegraphics[height = 0.13\linewidth]{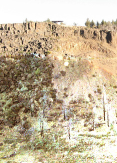} &
\includegraphics[height = 0.13\linewidth]{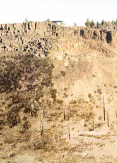} \\
(a) Input image & (b) Previous result & (c) Our result & (d) & (e) & (f)
\end{tabular}
\begin{tabular}{c@{\hspace{0.02in}}c@{\hspace{0.02in}}c@{\hspace{0.02in}}c@{\hspace{0.02in}}c@{\hspace{0.02in}}c}
\includegraphics[width = 0.15\linewidth]{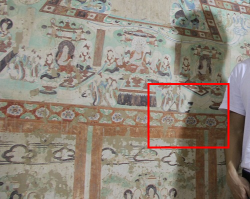} &
\includegraphics[width = 0.15\linewidth]{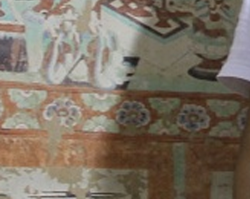} &
\includegraphics[width = 0.15\linewidth]{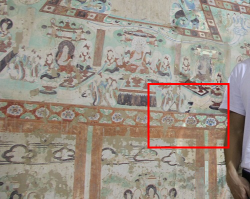} &
\includegraphics[width = 0.15\linewidth]{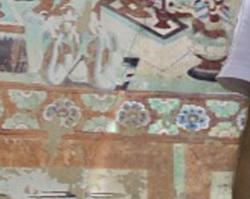} &
\includegraphics[width = 0.15\linewidth]{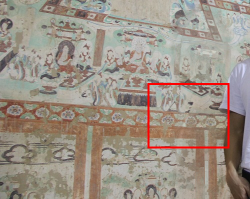} &
\includegraphics[width = 0.15\linewidth]{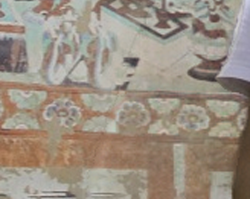} \\
\multicolumn{2}{c}{(g) Input image} & \multicolumn{2}{c}{(h) Previous result} & \multicolumn{2}{c}{(i) Our result} \\
\end{tabular}

    \caption{Top row: input image (a)(d); the state-of-the-art shadow removal method of Liu and Gliercher.~\protect\shortcite{eccv:2008} produces results with color inconsistencies (b)(e); our shadow region harmonization (SRH) method automatically corrects these issues (c)(f).
    Bottom row: input image (g); the state-of-the-art shadow removal method of Liu and Gliercher.~\protect\shortcite{Xiao132} produces results with texture inconsistencies (h); our shadow region harmonization (SRH) method automatically corrects these issues (i).}

    \label{fig:ia}

\end{figure*}

Cast shadows occur in many photography scenarios, and often lead to distracting artifacts that detract from the visual appeal of a photograph.
Removing cast shadows from such photographs is often highly desirable, yet difficult to achieve due to its inherently ill-posed nature: it is difficult for computational techniques, without any prior knowledge, to disambiguate shadows from dark textured regions in the scene. 



In the past decade, many approaches have been proposed for removing shadows in photographs. 
However, many of these techniques suffer from {\em inconsistency} artifacts, i.e. the de-shadowed region is visually incompatible with the rest of the input image. Most previous methods assume simplified shadow models that boil down to a simple color and intensity correction of the shadowed pixels. This assumption typically does not produce good results in presence of soft shadows, complex spatially-varying textures, complex reflectance properties of the underlying material (e.g., BRDF) and loss of dynamic range in the shadow region (see \figref{ia} top). Postprocessing, such as tone/color adjustment, gamma correction, lossy compression, can also easily violate common shadow models.


In this work we aim at providing new tools that can help users achieve high quality shadow removal results.
We propose a new technique called {\em Shadow Region Harmonization} (SRH), which can effectively remove inconsistency artifacts from existing shadow removal results.
Our method is built on the general idea of building correspondence
between shadow and non-shadow regions, and enforcing consistent color and texture properties of corresponding regions.
Our key insight is that the shadow region often (but not always) contains the same materials as in the non-shadow regions. Thus, correspondence between these two types of regions can be constructed {\em locally}, instead of globally. We build such correspondence using a guided patch-based image synthesis framework, where shadow regions are reconstructed using non-shadow ones. For each shadow patch, we use its corresponding non-shadow ones to compute a parametric appearance correction model based on color and texture.
To handle unique shadow materials, we compute for each patch a {\em correction confidence}, and use it in an optimization process to ensure that patches without good correspondences can also be corrected properly.
We quantitatively evaluate the SRH method on a recent benchmark dataset~\cite{GONG14}, and show that it can significantly improve the output of existing methods, including the state-of-the-art ones.

\section{Related Work}


We first briefly review representative works closely related to ours, and then discuss the inconsistency artifacts in previous methods in more detail.

\textbf{Shadow removal}
is an extensively studied problem and modern approaches are well summarized in recent surveys~\cite{Xu06shadowdetection,Sanin:2012}.
Shadow analysis is also closely related to intrinsic image decomposition~\cite{RIS78,grosse09intrinsic} -- the problem of separating an image into reflectance and illumination components --  though shadow removal focuses on the illumination variation caused by occluded light sources.


A common way to detect shadows is to use \emph{illuminant invariant features}~\cite{SFS14,02ECCV}, which help to detect shadow boundaries. Shadow removal can then be achieved by reconstructing the image with the shadow gradients edited~\cite{02ECCV,05Hami}.
%
%
Baba el al.~\shortcite{Baba:2004} estimate shadow density directly using patch lightness. Gryka et al.~\shortcite{Gryka2015softShadows} extract soft shadows by learning a regression function from image patches to shadow mattes. However, both these techniques use simple shadow models that are often inadequate for the non-linear images that are constitute the vast majority of photographs. Laplacian pyramid~\cite{CGF08} and gradient domain processing~\cite{eccv:2008} have been used to improve the consistency of textures between well-lit and shadowed regions. These techniques have limitations when there are multiple textures in the same shadow region. Guo et al.~\shortcite{11cvpr} build a classifier to detect shadowed and non-shadowed region pair of the same texture, however it is based on image segmentation, which itself could be fragile on complex scenes.

Because of the inherently ambiguous nature of shadow detection and removal, many previous approaches require manual specification of the shadow region~\cite{CGF08,eccv:2008,ESS:07}. Given this input, shadow removal can be posed as a matting~\shortcite{tog07} or labeling~\shortcite{ISR:09} problem.
%
%


\textbf{Patch-based synthesis} has shown great success in image and video completion~\shortcite{Wexler:2007:SCV} and other editing tasks since the introduction of \emph{PatchMatch}~\cite{Barnes:2009:PatchMatch} -- a fast approximate method for computing patch-based dense correspondences. PatchMatch has been generalized to support scaling and rotation of patches~\cite{Barnes:2010:GPatchMatch} as well as gain/bias of each individual color channels~\cite{HaCohen:2011:NRDC}.
This family of techniques have been widely used for finding patch correspondence in a rotation and scale-invariant manner, and can also handle differences in illumination conditions.

Our SRH algorithm uses guided patch-based synthesis~\cite{Wexler:2007:SCV} to reconstruct the shadow region using non-shadow patches, using the initial shadow removal result as guidance.
This is similar to the Image Analogies framework~\cite{Hertzmann:2001} that was used for guided texture synthesis and image enhancement.  
In our method we address a unique challenge: the shadow region may contain unique structures/materials, and thus patch synthesis can only be partially successful. We use an optimization approach that gracefully combines traditional shadow color correction with patch-based synthesis to generate consistent removal results in the entire shadow region. Interestingly, Gryka et al.~\shortcite{Gryka2015softShadows} also used patch-based image completion~\cite{Wexler:2007:SCV} to compute one of the features for its learning based framework. However the synthesis was not guided so the completed content might end up looking very different than the real one. 

\textbf{Inconsistency artifacts of shadow removal}
Existing shadow removal approaches often produce inconsistency artifacts in recovered shadow regions, due to the violation of the simplified shadow models they use; these include both color and texture inconsistencies.
Specifically, most approaches cannot model the loss of dynamic range in shadow regions~\cite{eccv:2004,ESS:07}, which leads to inconsistent noise properties and texture characteristics between recovered shadow regions and non-shadow regions, such as the examples in \figref{res2}.
Pixel-based approaches~\cite{02ECCV,eccv:2004,Baba:2004} suffer from inaccuracies in the estimation of the shadow parameters, leading to color shifts or residual shadows in the recovered shadow regions(\figref{ia}(top)). To correct such artifacts some methods leverage shadow/non-shadow region correspondence~\cite{CGF08,eccv:2008,GONG14}, or region-based color transfer~\cite{ICME2013}. However, they are still not robust to complex spatially-varying textures, complex reflection and shading properties.
\figref{res1}(e) shows an example with a colorful translucent occluder, where the green color can not be eliminated using any existing shadow removal methods.

Our SRH approach builds dense correspondence between shadow and non-shadow regions to enforce both color and texture consistency,
as shown in \figref{ia}.
It is an automatic post-processing method that is independent of specific shadow removal approach being applied first, and thus can be applied widely.

\begin{figure*}[t]
    \centering
\begin{tabular}{c@{\hspace{0.02in}}c@{\hspace{0.02in}}c@{\hspace{0.02in}}c@{\hspace{0.02in}}c@{\hspace{0.02in}}c}
\includegraphics[width = 0.16\linewidth]{images/sec_new_algo/2/1.png} &
\includegraphics[width = 0.16\linewidth]{images/sec_new_algo/2/1_n.png} &
\includegraphics[width = 0.16\linewidth]{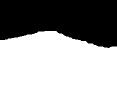} &
\includegraphics[width = 0.16\linewidth]{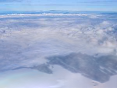} &
\includegraphics[width = 0.16\linewidth]{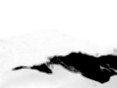} &
\includegraphics[width = 0.16\linewidth]{images/sec_new_algo/2/1_cor_res.png} \\
\includegraphics[width = 0.16\linewidth]{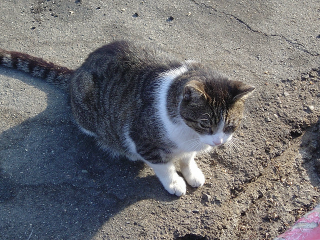} &
\includegraphics[width = 0.16\linewidth]{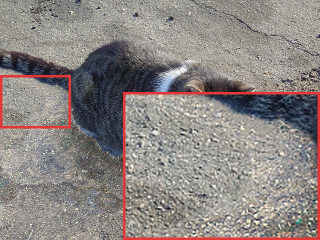} &
\includegraphics[width = 0.16\linewidth]{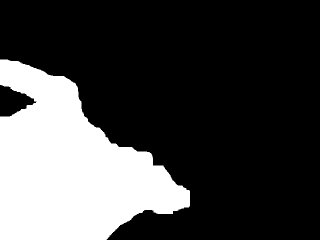} &
\includegraphics[width = 0.16\linewidth]{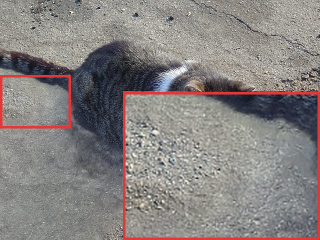} &
\includegraphics[width = 0.16\linewidth]{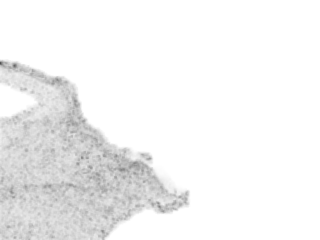} &
\includegraphics[width = 0.16\linewidth]{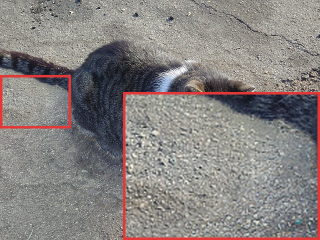} \\
(a) Input image & (b) Initial result & (c) Shadow mask & (d) Patch synthesis, $I^S$ & (e) Confidence, $C$ & (f) Final result, $I^C$
\end{tabular}
    \caption{Given an input image $I$ (a), we compute an initial shadow removal result, $I^N$ (b) using current shadow removal methods (top, \protect\cite{eccv:2008}, bottom, \protect\cite{ISR:09}). We use this result to estimate a shadow mask (c), and a patch-based synthesized image, $I^S$ (d). This result is inaccurate in regions that are not observed in the non-shadow region and this is captured by the confidence map, $\boldsymbol{C}$ (e). Combining $I^C$ and $I^S$ using $C$ gives us the final correction, $I^C$, (f) where the inconsistency artifacts in $I^N$ are removed.}
    \label{fig:cmgillu}
\end{figure*}

\section{Our Approach}

Our SRH approach takes the original image, $I$, and an initial shadow removal result, $I^N$, as input, and produces a higher quality result, $I^H$.

\subsection{The Harmonization Model} \label{sec:model}

In contrast to previous shadow removal methods that use pixelwise shadow models, the SRH method is based on image patches that capture better local color and texture characteristics.
It computes a parametric {\em Appearance Harmonization Model}, which describes for each shadow patch, how to change its color and texture to make it more consistent with its corresponding non-shadow patches.
It contains three components: (1) color correction parameters, (2) texture correction parameters, and  (3) a correction confidence.

\textbf{Color Correction}
We model the effect of a shadow as a per-channel affine transform on the pixel values. Shadows tend to be spatially smooth, and we incorporate this prior by assuming that this affine transform is constant within a small patch (e.g. $5 \times 5$). This gives us the following color correction model:
\begin{equation} \label{equ:patchconsist}
I^{H}_{c}(P) = g^{c}(P) \cdot I^N_{c}(P) + b^{c}(P),
\end{equation}
where $c$ is the color channel index,  $I^{H}(P)$ and $I^N(P)$ correspond to patches $P$ in $I^H$ and $I^N$, and $g^{c}(P)$ and $ b^{c}(P)$ are per-patch, per-channel gain and bias values that together constitute our affine color correction term.



\textbf{Texture Correction}
The standard deviation of color channels within an image patch is often used to describe local texture and image details~\cite{CGF08,eccv:2008}.
We denote the standard deviation of patch $P$ and color channel $c$ in $I^N$ and $I^H$ as $\sigma^N_c(P)$ and $\sigma^H_c(P)$, respectively. We scale $\sigma^N(P)$ as:
\begin{equation} \label{equ:textureconsist}
 \sigma^H_c(P) = s_c(P)\cdot \sigma^N_c(P),
\end{equation}
so that shadow patches have the same level of local color contrast as the corresponding non-shadow ones.
Note that the scale $s_c$ is different from the gain in \equref{patchconsist} as it controls the deviation around the mean color (vs. 0).


\textbf{Correction Confidence}
To estimate the above correction parameters for each patch, we use a patch synthesis approach to match shadow and non-shadow patches of the same material. Due to limitations of patch synthesis, not all matches are reliable, and the estimated parameters at these patches are incorrect.
To describe the reliability of the correction parameters of a specific shadow patch $P$, we compute an extra confidence value $C(P)$ in $[0,1]$ and add it to the parameter list.
We will describe how to compute this value in \secref{corremap}.

\textbf{Final Model}
Allowing gain, bias and texture scaling in every color channel will result in a 10 dimensional parameter vector for each pixel. To maintain a good balance between model complexity and flexibility,
in practice we choose to apply corrections in CIELab space, and only enable gain in the L channel, bias in the $a$ and $b$ channels for color correction, and only the L channel for texture correction.
This gives us a 5-channel parameter map $\boldsymbol{S} = (\boldsymbol{g}^L, \boldsymbol{b}^A, \boldsymbol{b}^B, \boldsymbol{s}^L, \boldsymbol{C})$,
which we refer to as the \emph{shadow correction map}.
Other parameter combinations will be discussed and compared in \secref{cp}.


\subsection{Shadow Correction Map Generation} \label{sec:cmg}

We now describe how to estimate the shadow correction map $S$, given the source image $I$ and an initial shadow removal result $I^N$. Our key idea is to build dense correspondences between shadow and non-shadow patches, and derive the correction parameters from them in a way that unique shadow structures and materials can be properly handled.

A binary mask is needed to indicate which pixels are inside the shadow region that need to be corrected. In our work we assume the only difference between $I^N$ and $I$ is the color of shadow pixels. We apply a small threshold on the difference image $I^N-I$ to generate the correction region $\mathcal{R}_{c}$, and further apply a small dilation operation using a $3\times 3$ kernel to remove occasional small holes inside it. This is shown in \figref{cmgillu}(c).

\subsubsection{Patch-based Synthesis} \label{sec:pbs}

We use patch-based synthesis to synthesize a new shadow-free image, $I^S$, where the correction region $\mathcal{R}_{c}$ has been filled using patches from the source region $\mathcal{R}_{s}=I\backslash \mathcal{R}_{c}$.
We use a guided variant of the \emph{Image Melding} algorithm~\cite{Darabi12:ImageMelding12} as the basis for this synthesis task, given its support for patch scaling, rotation, reflection, and color gain and bias.
It is applied in a coarse-to-fine manner on an image pyramid.

Despite its color and texture inconsistencies, $I^N$ usually gives us a good initial estimation of the final result that can be used as guidance for the synthesis process.
We thus use $I^N$ in two ways. First, we use it to initialize the synthesis result, $I^S$, at the coarsest level. Second, in the spirit of Image Analogies~\cite{Hertzmann:2001}, we use it as a guidance layer during the synthesis process.
Specifically, the distance between a patch $P$ in $\mathcal{R}_{c}$ and a patch $Q$ in $\mathcal{R}_{s}$ is defined as:
\begin{align}\label{equ:syn}
&d(P, Q) \;=\; || I^S(P), I^S(Q) ||_2 \; + \nonumber \\
				     &\beta \; || I^N(P) \cdot \boldsymbol{g}(P) + \boldsymbol{b}(P), I^S(P)||_2 + \gamma \; \mathcal{E}(\boldsymbol{g}(P), \boldsymbol{b}(P)),
\end{align}
where $|| I(P), I(Q) ||_2$ is the average $L2$ color distance of two patches, and $\beta$ and $\gamma$ are balancing weights.
The second term is our guidance term; it constrains the synthesized patch $I^S(P)$ to be similar to the initial shadow removal result $I^N(P)$.
The additional gain $\boldsymbol{g}(P)$ and bias $\boldsymbol{b}(P)$  are introduced in this term to compensate for the possible color and intensity inconsistencies in $I^N$. Since we expect $I^N$ to be reasonably close to the final result, we also add a third term
$\mathcal{E}(\boldsymbol{g}(P_i), \boldsymbol{b}(P_i))$ to punish unrealistically large gain and bias, defined as ($P$ is omitted):
\begin{equation}
\mathcal{E}(\boldsymbol{g}, \boldsymbol{b}) = \sum_c (| \boldsymbol{g}^c - 1 | + | \boldsymbol{b}^c - 0 |), c \in \{L, A, B\}.
\end{equation}
We use this patch synthesis process produces to reconstruct $I^S$, which will be used in the next step to derive the correction map.
\figref{cmgillu}(d) shows two synthesis results.

\textbf{Parameter settings } We decompose the input image into a pyramid with a coarsest scale of size 30 pixels (smaller dimension), and increase the scale by a factor of $1.4$ for each pyramid level. $\beta$ is set to $30$. Patch distance is computed in CIELab space, and gain ($L$ channel), bias ($a$ \& $b$ channels) ranges are set to $[0.9, 1.11]$, $[-0.05, 0.05]$, respectively.
$\gamma$ is set to $4$.

\subsubsection{Computing Correction Parameters}\label{sec:corremap}

The synthesis result $I^S$ can not be directly used as the final output for several reasons.
Firstly, patch synthesis does not perform well for regions with unique structures, as shown in \figref{cmgillu}(d)(top).
Secondly, patch synthesis may not converge well especially for highly textured regions, resulting in blurry results, as shown in the example in \figref{cmgillu}(d)(top and bottom).
Nevertheless, $I^S$ contains good color and texture information in a large portion of the shadow region, that can be used to estimate the corresponding parameters in the correction map.

\textbf{Color Parameters}
For each patch $P$ in the correction region, the color correction parameters can be derived from \equref{patchconsist} as:
\begin{eqnarray}
\boldsymbol{g}^L(P) &=& \sum_{p \in P} I^S_L(p) / \sum_{p \in P}I^N_L(p), \nonumber \\
\boldsymbol{b}^A(P) &=& \frac{1}{||P||} (\sum_{p \in P} I^S_A(p) - \sum_{p \in P}I^N_A(p)), \nonumber \\
\boldsymbol{b}^B(P) &=& \frac{1}{||P||}(\sum_{p \in P} I^S_B(p) - \sum_{p \in P}I^N_B(p)),
\label{equ:con1}
\end{eqnarray}
where $L$, $A$, $B$ denote the luminance and chrominance channels.

\textbf{Texture Parameters}
Directly computing the texture correction parameter from the synthesis result $I^S$ is sub-optimal given that it may be blurry and lack image details ( see \figref{cmgillu}(d)(bottom)).
We instead directly use the patch correspondence, instead of final synthesis result, to estimate this parameter.
That is, for patch $P$ in $\mathcal{R}_{c}$, we get a source region patch $Q$ that is used to vote for the final synthesis result.
We assume $P$ should have the same texture characteristics as $Q$,
So the texture consistency parameter can be computed as:
\begin{equation}\label{equ:textureparam}
\boldsymbol{s}^L(P) = \sigma_Q / \sigma_P.
\end{equation}

\textbf{Correction Confidence}
Given the original shadow removal result $I^N$ and the patch synthesis result $I^S$, the synthesis confidence $\boldsymbol{C}(P)$ for each patch $P$ is defined as:
\begin{equation} \label{equ:corr_confidence}
\boldsymbol{C}(P) = 1 - \frac{min_{\mathcal{\boldsymbol{g}, \boldsymbol{b}}} || I^N(P) * \boldsymbol{g}(P) + \boldsymbol{b}(P) - I^S(P)||_2 } { || I^N(P) ||_2 + \varepsilon },
\end{equation}
%
where the distance between $I^N(P)$ and $I^S(P)$ is minimized by searching the best gain for the $L$ channel, and the best bias for the $a$ and $b$ channels per patch.
The confidence is normalized by the average pixel luminance $|| I^N(P) ||_2$ to avoid a bias in dark regions. $\varepsilon$ is a small constant to avoid division by zero.
Intuitively, if $I^N(P_i)$ and $I^S(P_i)$ contain the same structure and only differ by a global color transform, we have high confidence that the patch synthesis result is correct, and the correction parameters is reliable.
Otherwise if $I^N(P_i)$ and $I^S(P_i)$ contain structural differences, the confidence value will be low, indicating incorrect patch synthesis.
Example confidence maps are shown in \figref{cmgillu}(e); note that
unique structures in the shadow region such as the dark brown mountain in \figref{cmgillu}(top) cannot be synthesized well, and consequently pixels inside these structures have very small confidence values.

Next, we describe how to use the confidence map to refine the parameters in the correction map.

\subsubsection{Correction Map Refinement} \label{sec:refine}

The initial correction map is not reliable for all shadow patches, and thus cannot be directly applied to $I^N$.
In this section, we show how to refine it based on the computed correction confidence values (\equref{corr_confidence}).
We treat color and texture parameters differently at this stage due to their inherently different nature.

\textbf{Color Parameter Refinement}
Experiments show that the color correction parameters (gain/bias of CIELAB channel) are generally quite smooth in the shadow regions.
 To refine color parameters, we propagate them from high confidence patches to low confidence ones, by optimizing a quadratic objective function.
Specifically, for channel $k$ ($k={L,A,B}$) of the original color correction parameters, denoted by $S_0^k$, we find new parameters $S^k$ that minimize the following energy function:
\begin{align}
& E^k = \sum_{P} \boldsymbol{C}(P) \cdot (S^k(P) - S_0^k(P))^2 \nonumber \\
      & + \lambda_s \sum _{ Q \in \mathcal{N}(P) } (1- \boldsymbol{C}(P))\cdot \mathcal{A}(P, Q) \cdot(S^k(P) - S^k(Q))^2,
\end{align}
where $\mathcal{N}(P)$ is the set of the neighboring overlapping patches of $P$.
The first term is the data term that constrains $S^k$ to be close to the original estimation $S^k_0$ if the synthesis confidence $C(P)$ is high.
The second term is the smoothness term, weighted by the color difference of neighboring patches, along with $1-C$.
The affinity weight $\mathcal{A}(P, Q)$ is defined as:
\begin{equation}
\mathcal{A}(P, Q) = \mathcal{G}( || I_n(P) - I_n(Q) ||_2, \sigma_m),
\end{equation}
where $\mathcal{G}(x, \sigma) = exp(- x^2 / \sigma^2)$ is the Gaussian function.
The smooth term allows low confidence patches to receive parameter values from neighboring patches that have similar colors to them. In our implementation we set $\lambda_s = 10$ and $\sigma_m = 0.2$, and solve the linear system using Conjugate Gradient.

\textbf{Texture Parameter Refinement}
Unlike the color parameters, texture parameters are very noisy over the whole image. 
We thus cannot refine them using a similar optimization.
Instead for patches with low correction confidence, we resort to the initial shadow removal result for computing the texture parameters. In other words, for regions where patch synthesis is not reliable, we maintain their original texture characteristics in the initial shadow removal result, to avoid introducing additional artifacts.
Specifically, we use the correction confidence as the interpolation coefficient to compute the refined texture parameter $s^L(P)$ as:
\begin{equation}
s^L(P) = \boldsymbol{C}(P) \cdot s^L_0(P) + (1- \boldsymbol{C}(P)) \cdot 1,
\end{equation}
Where $s^L_0(P)$ is the computed texture parameter using \equref{textureparam} on the  synthesized image.


\begin{figure}[tb]
    \centering
\begin{tabular}{c@{\hspace{0.05in}}c@{\hspace{0.05in}}c@{\hspace{0.05in}}c}
\begin{sideways} \hspace{0.1in} (a) \end{sideways} &
\includegraphics[width = 0.3\linewidth]{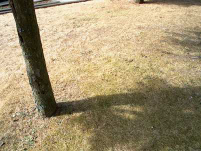} &
\includegraphics[width = 0.3\linewidth]{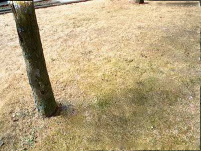} &
\includegraphics[width = 0.3\linewidth]{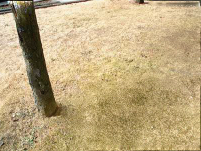} \\
\begin{sideways} \hspace{0.1in} (b) \end{sideways} &
\includegraphics[width = 0.3\linewidth]{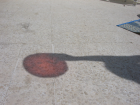} &
\includegraphics[width = 0.3\linewidth]{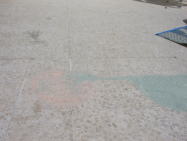} &
\includegraphics[width = 0.3\linewidth]{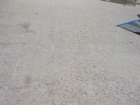} \\
\begin{sideways} \hspace{0.1in} (c) \end{sideways} &
\includegraphics[width = 0.3\linewidth]{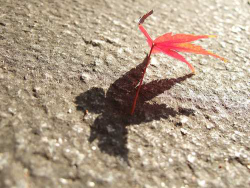} &
\includegraphics[width = 0.3\linewidth]{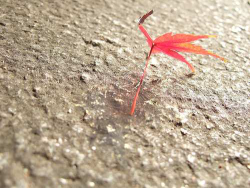} &
\includegraphics[width = 0.3\linewidth]{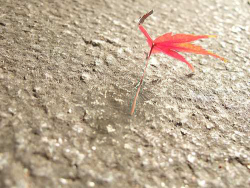} \\
\begin{sideways} \hspace{0.1in} (d) \end{sideways} &
\includegraphics[width = 0.3\linewidth]{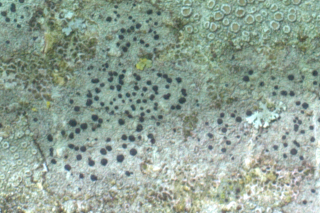} &
\includegraphics[width = 0.3\linewidth]{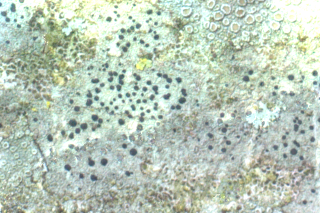} &
\includegraphics[width = 0.3\linewidth]{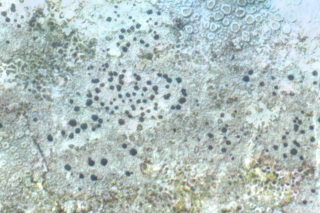} \\
\begin{sideways} \hspace{0.1in} (e) \end{sideways} &
\includegraphics[width = 0.3\linewidth]{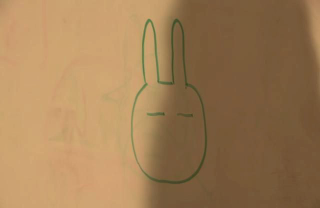} &
\includegraphics[width = 0.3\linewidth]{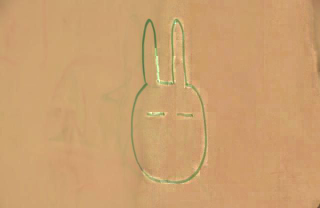} &
\includegraphics[width = 0.3\linewidth]{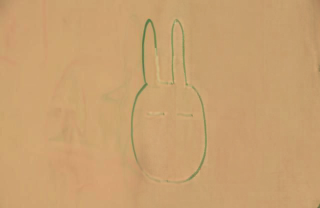} \\
& Source image & Previous result $I^N$ & Our result $I^C$\\
\end{tabular}
    \caption{Improving previous shadow removal results with color inconsistency using the proposed SRH algorithm. (a) \protect\cite{eccv:2004}, (b) \protect\cite{CGF08}, (c) \protect\cite{ICME2013}, (d) \protect\cite{GONG14}, (e) \protect\cite{Gryka2015softShadows}. }
    \label{fig:res1}
\end{figure}

%
%
%
%

\subsection{Applying the Correction Map} \label{sec:ofr}

The final recovery result is obtained by
 applying the shadow correction map $S$ on the initial shadow removal result $I^N$.
Specifically, to remove color inconsistency, we apply color correction parameters $(\boldsymbol{g}^L, \boldsymbol{b}^A, \boldsymbol{b}^B)$ of $S$ to each patch and vote for the output image $I^D$ as:
\begin{eqnarray}
I^D_L(p) = \frac{1}{||P||} \sum_{q \in P} I^N_L(q) * \boldsymbol{g}^L(Q), \nonumber \\
I^D_A(p) = \frac{1}{||P||} \sum_{q \in P} (I^N_A(q) + \boldsymbol{b}^A(Q)), \nonumber \\
I^D_B(p) = \frac{1}{||P||} \sum_{q \in P} (I^N_B(q) + \boldsymbol{b}^B(Q)),
\label{equ:con2}
\end{eqnarray}
where $P$ and $Q$ are patches centered at pixel $p$ and $q$, respectively.
Furthermore, to remove texture inconsistency, we apply texture correction parameters  $\boldsymbol{s}^L$ of each patch and vote for the final recovery result $I^C$:
\begin{equation}
I^C_L(p) = \frac{1}{||P||} \sum_{p \in Q} (\frac{\boldsymbol{s}^L(Q) \sigma^N_L(Q)}{\sigma^D_L(Q)}(I^D_L(p) - \mu_L^D(Q)) + \mu_L^D(Q)),
\label{equ:con2}
\end{equation}
where $\mu^D(Q)$, $\sigma^D(Q)$ are the average color and standard derivation of patch $Q$ in $I^D$, and $\sigma^N_L(Q)$ is the standard derivation of patch $Q$ in $I^N$.

\figref{cmgillu}(f) shows examples of the corrected shadow regions.
We can see that the color and texture inconsistency in the initial shadow removal results have been highly suppressed, resulting in more natural shadow removal results.

\begin{figure*}[htb]
\setlength{\tabcolsep}{1pt}
    \centering
\begin{tabular}{c@{\hspace{0.05in}}c@{\hspace{0.05in}}c@{\hspace{0.12in}}c@{\hspace{0.05in}}c@{\hspace{0.12in}}c@{\hspace{0.05in}}c}
\begin{sideways} \hspace{0.3in} (a)  \end{sideways} &
\includegraphics[width = 0.15\linewidth]{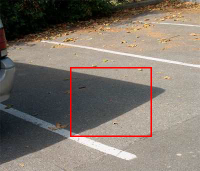} &
\includegraphics[width = 0.15\linewidth]{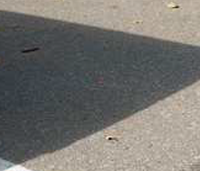} &
\includegraphics[width = 0.15\linewidth]{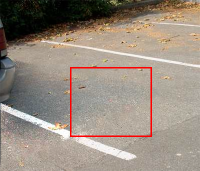} &
\includegraphics[width = 0.15\linewidth]{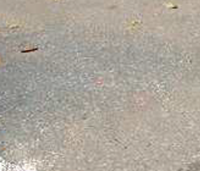} &
\includegraphics[width = 0.15\linewidth]{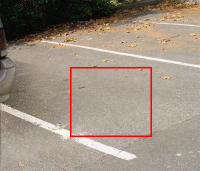} &
\includegraphics[width = 0.15\linewidth]{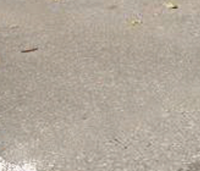} \\
\begin{sideways} \hspace{0.3in} (b)  \end{sideways} &
\includegraphics[width = 0.15\linewidth]{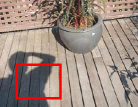} &
\includegraphics[width = 0.15\linewidth]{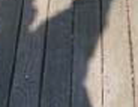} &
\includegraphics[width = 0.15\linewidth]{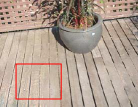} &
\includegraphics[width = 0.15\linewidth]{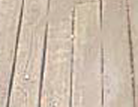} &
\includegraphics[width = 0.15\linewidth]{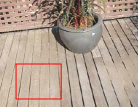} &
\includegraphics[width = 0.15\linewidth]{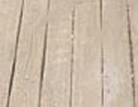} \\
\begin{sideways} \hspace{0.35in} (c)  \end{sideways} &
\includegraphics[width = 0.15\linewidth]{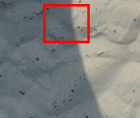} &
\includegraphics[width = 0.15\linewidth]{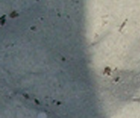} &
\includegraphics[width = 0.15\linewidth]{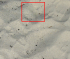} &
\includegraphics[width = 0.15\linewidth]{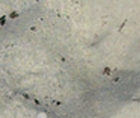} &
\includegraphics[width = 0.15\linewidth]{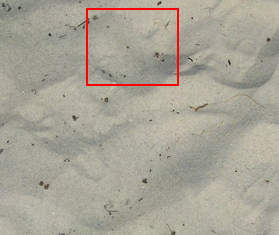} &
\includegraphics[width = 0.15\linewidth]{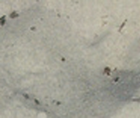} \\
\begin{sideways} \hspace{0.25in} (d)  \end{sideways} &
\includegraphics[width = 0.15\linewidth]{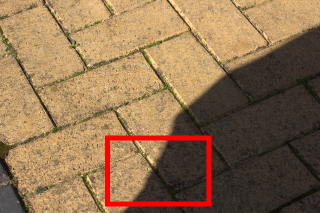} &
\includegraphics[width = 0.15\linewidth]{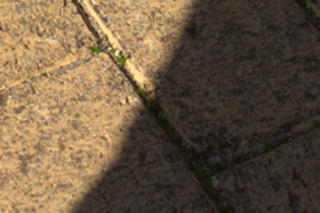} &
\includegraphics[width = 0.15\linewidth]{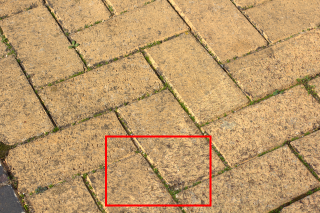} &
\includegraphics[width = 0.15\linewidth]{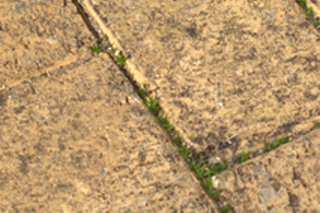} &
\includegraphics[width = 0.15\linewidth]{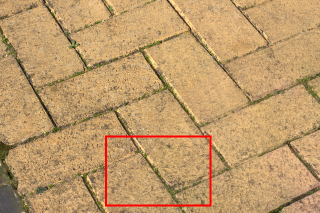} &
\includegraphics[width = 0.15\linewidth]{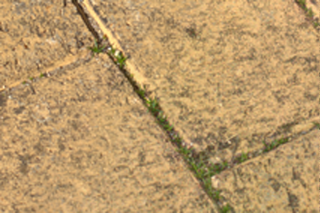} \\
\begin{sideways} \hspace{0.25in} (e)  \end{sideways} &
\includegraphics[width = 0.15\linewidth]{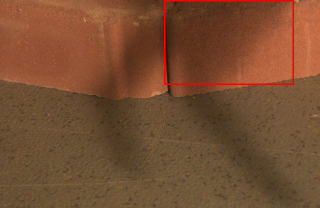} &
\includegraphics[width = 0.15\linewidth]{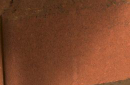} &
\includegraphics[width = 0.15\linewidth]{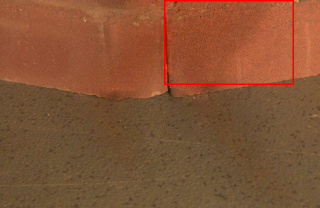} &
\includegraphics[width = 0.15\linewidth]{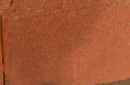} &
\includegraphics[width = 0.15\linewidth]{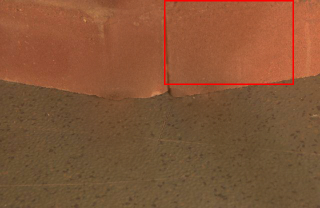} &
\includegraphics[width = 0.15\linewidth]{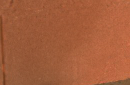} \\
& \multicolumn{2}{c}{Input image} & \multicolumn{2}{c}{Previous result $I^N$} & \multicolumn{2}{c}{Our result $I^C$} \\
\end{tabular}
    \caption{Improving previous results with texture inconsistency using SRH.  (a) \protect\cite{eccv:2004}, (b) \protect\cite{05Bys}, (c) \protect\cite{ICME2013}, (d) \protect\cite{GONG14}, (e) \protect\cite{Gryka2015softShadows}.  }
    \label{fig:res2}
\end{figure*}

\begin{figure*}[htb]
		\setlength{\tabcolsep}{1pt}
		\centering
\begin{tabular}{c@{\hspace{0.02in}}c@{\hspace{0.02in}}c@{\hspace{0.02in}}c@{\hspace{0.02in}}c@{\hspace{0.02in}}c@{\hspace{0.02in}}c}
\begin{sideways} \hspace{0.04in} Input images  \end{sideways} &
\includegraphics[width = 0.15\linewidth]{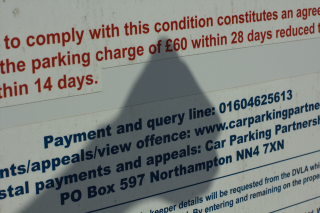} &
\includegraphics[width = 0.15\linewidth]{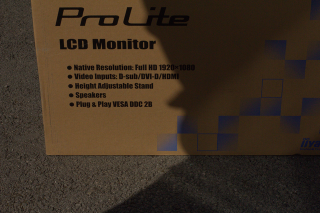} &
\includegraphics[width = 0.15\linewidth]{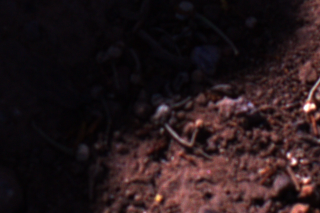} &
\includegraphics[width = 0.15\linewidth]{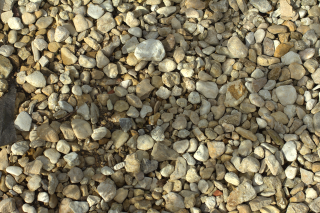} &
\includegraphics[width = 0.15\linewidth]{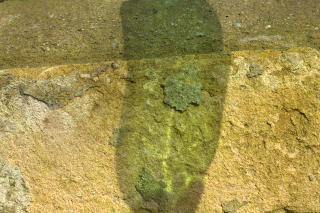} &
\includegraphics[width = 0.15\linewidth]{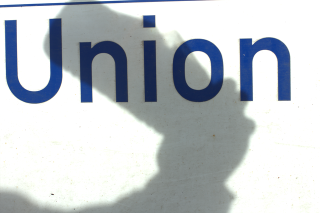} \\
\begin{sideways} \hspace{0.05in} Initial result  \end{sideways} &
\includegraphics[width = 0.15\linewidth]{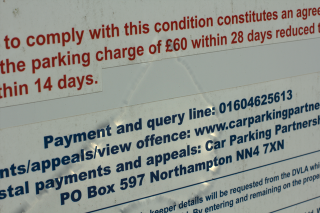} &
\includegraphics[width = 0.15\linewidth]{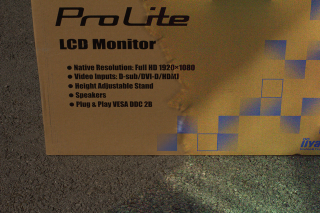} &
\includegraphics[width = 0.15\linewidth]{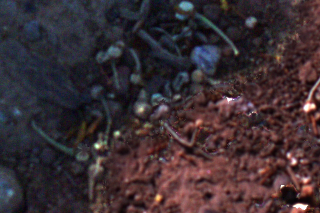} &
\includegraphics[width = 0.15\linewidth]{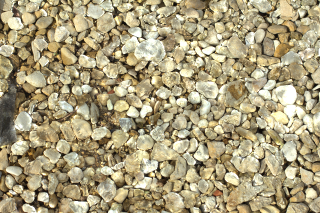} &
\includegraphics[width = 0.15\linewidth]{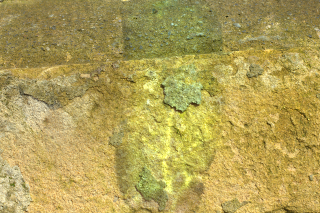} &
\includegraphics[width = 0.15\linewidth]{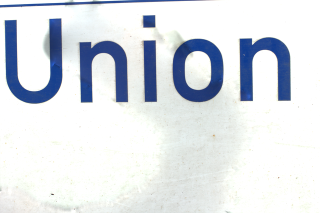} \\
\begin{sideways} \hspace{0.10in} Our result  \end{sideways} &
\includegraphics[width = 0.15\linewidth]{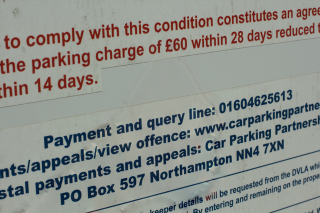} &
\includegraphics[width = 0.15\linewidth]{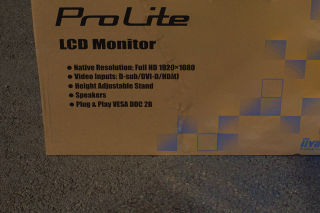} &
\includegraphics[width = 0.15\linewidth]{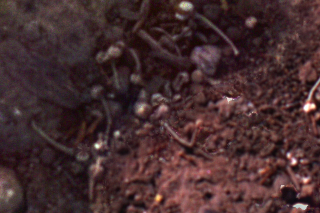} &
\includegraphics[width = 0.15\linewidth]{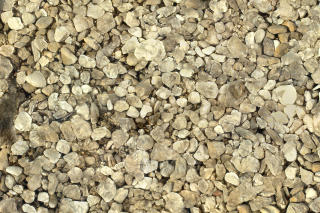} &
\includegraphics[width = 0.15\linewidth]{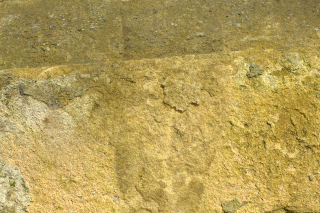} &
\includegraphics[width = 0.15\linewidth]{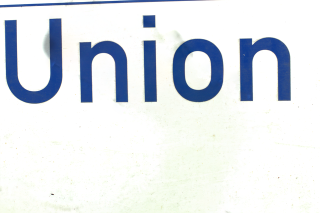} \\
\end{tabular}
    \caption{Improving state-of-the-art results on examples in the benchmark dataset \protect\cite{GONG14}. (top) Original input images; (middle) Initial shadow removal results $I^N$ using \protect\cite{GONG14}. (bottom) Our harmonized result $I^C$.}
    \label{fig:res3}
\end{figure*}

\begin{figure*}[htb]
    \centering
\begin{tabular}{c@{\hspace{0.07in}}c@{\hspace{0.02in}}c@{\hspace{0.07in}}c@{\hspace{0.02in}}c@{\hspace{0.07in}}c@{\hspace{0.02in}}c}
\includegraphics[width = 0.07\linewidth]{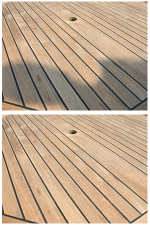} &
\includegraphics[width = 0.14\linewidth]{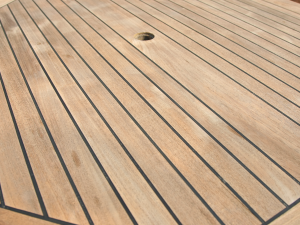} &
\includegraphics[width = 0.14\linewidth]{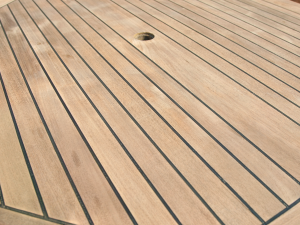} &
\includegraphics[width = 0.14\linewidth]{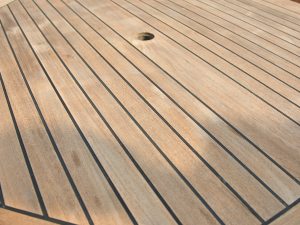} &
\includegraphics[width = 0.14\linewidth]{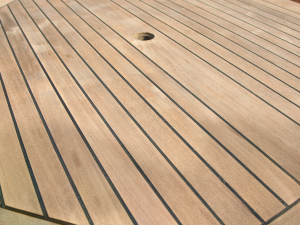} &
\includegraphics[width = 0.14\linewidth]{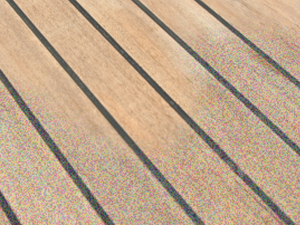} &
\includegraphics[width = 0.14\linewidth]{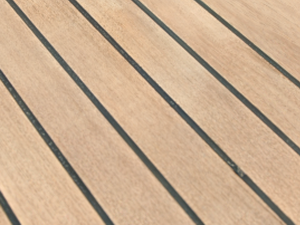} \\
(a) Src & (b) Syn. $I^N$($\alpha = 0.2$) & (c) Our result & (d) Syn. $I^N$($\alpha = 0.5$) & (e) Our result & (f) Adding noise & (g) Our result \\
 images & (error = 0.034) & (error = 0.027) & (error = 0.086) & (error = 0.059) & (error = 0.066) & (error = 0.029)\\
\end{tabular}
    \caption{Results generated by our SRH method with synthetic $I^N$. (a) Full-shadow image \& Shadow-free image; (b) and (d) Synthetic $I^N$ generated by linear interpolation of full-shadow image and shadow-free image with $\alpha = 0.2, 0.5$; (c) and (e) SRH result applied on (b) and (d); Our algorithm is robust to an initial results with 20\% residual shadow. (f) Close-up of a synthetic $I^N$ with 20\% residual shadow and strong Gaussian noise; (g) SRH result applied on (f). Our algorithm is robust to the apparent noise. Average errors w.r.t ground truth are reported in the bottom.
		}
    \label{fig:qr}
\end{figure*}

\section{Results and Evaluations}

We have implemented our algorithm in C++. On a PC with 3.4GHz CPU and 2G RAM, for a $600\times450$ image, our single-threaded implementation of the SRH algorithm takes about 2 minutes for the patch-based synthesis step, and 2 seconds for the rest - correction map construction, refinement and obtaining final result.
In this section, we evaluate the SRH method by showing both visual examples and quantitative evaluation results on a benchmark dataset.

\subsection{Visual Comparisons}

Figs.~\ref{fig:ia}(top), and \ref{fig:res1}, show shadow removal results from previous methods that contain significant color inconsistencies in the recovered shadow regions, including color shifts (\figref{ia}(top), \figref{res1}((b)(d)) and residual shadows (\figref{res1}(a)(c)).
We have experimented with a wide range of methods include \cite{eccv:2004,CGF08,eccv:2008,ICME2013,GONG14,Gryka2015softShadows}. The inputs for our SRH algorithm and the results of these methods are all taken from their original papers~\footnote{Ideally we should have compared all the methods against a common set of examples. However, we did not have access to the code of many of these methods. Instead, we compare against each method on its own successful examples reported in the original paper.}.
SRH successfully removes the color inconsistencies in the original results, and produces results that are more natural-looking.

In \figref{ia}(bottom), \figref{cmgillu}(bottom) and \figref{res2} we show shadow removal results on previous methods that contain texture inconsistencies, such as inconsistent noise properties and texture contrast.
These methods include \cite{eccv:2004,05Bys,ICME2013,Xiao132,GONG14}. Again, our SRH method successfully corrects these texture artifacts and produces more consistent shadow removal results.

In \figref{res3} we show the results of some algorithms on images from a recently proposed shadow removal benchmark dataset~\cite{GONG14}, along with improved results using our method. Again, the SRH method successfully suppresses both color and texture inconsistencies.



\subsection{Benchmark Evaluation}

We comprehensively evaluate our algorithm on the benchmark dataset mentioned above. This dataset consists of 214 test images, and provides quantitative errors of shadow removal results according to four attributes (more details in~\cite{GONG14}): texture, brokenness, colorfulness and softness.
The authors have also published shadow removal results using their technique as well as two other algorithms~\cite{11cvpr,ICME2013} for all 214 test examples.
We apply SRH on all the test cases using their shadow removal results as input, and report new errors and improvements in \tabref{perform}.
Due to limited space, we only show the average score of each attribute.

The results show that our SRH method reduces shadow removal errors for all categories and all the three previous methods.
Note that our algorithm cannot improve cases with detction errors, where the shadow region is wrongly detected (more in \secref{lim}).
Given that all these approaches introduce detection errors in some test cases, the performance improvement on examples with well-detected shadows are even higher.
Some visual comparisons are shown in \figref{res3}.

\begin{table*}[tb]
\centering
\small
\begin{tabular} {|c|c|c|c|c|c|c|}
\hline
& \multicolumn{2}{c|}{Gong \protect\shortcite{GONG14}} & \multicolumn{2}{c|}{Gong \protect\shortcite{ICME2013}} & \multicolumn{2}{c|}{Guo \protect\shortcite{11cvpr}} \\ \cline{2-7}
& $E_r$ & $E_r*$ & $E_r$ & $E_r*$ & $E_r$ & $E_r*$\\ \hline
Tex. & 0.34 : 0.32 (\textbf{6\%}) & 0.21 : 0.17 (\textbf{19\%}) & 0.47 : 0.43 (\textbf{9\%}) & 0.36 : 0.31 (\textbf{14\%}) & 0.61 : 0.56 (\textbf{8\%}) & 0.51 : 0.44 (\textbf{14\%}) \\ \hline
Soft. & 0.40 : 0.37 (\textbf{8\%})  & 0.23 : 0.22 (\textbf{12\%}) & 0.51 : 0.45 (\textbf{12\%}) & 0.40 : 0.32 (\textbf{20\%}) & 0.77 : 0.69 (\textbf{10\%}) & 0.68 : 0.56 (\textbf{18\%}) \\ \hline
Bro. & 0.42 : 0.40 (\textbf{5\%}) & 0.25 : 0.22 (\textbf{12\%}) & 0.59 : 0.53 (\textbf{10\%}) & 0.52 : 0.41 (\textbf{21\%}) & 0.81 : 0.77 (\textbf{5\%}) & 0.76 : 0.70 (\textbf{8\%}) \\ \hline
Col. & 0.44 : 0.40 (\textbf{9\%})  & 0.29 : 0.23 (\textbf{21\%}) & 0.75 : 0.72 (\textbf{4\%}) & 0.69 : 0.65 (\textbf{6\%}) & 1.12 : 1.01 (\textbf{10\%}) & 1.09 : 0.93 (\textbf{15\%}) \\ \hline
Other & 0.40 : 0.36 (\textbf{10\%}) & 0.26 : 0.21 (\textbf{19\%}) & 0.57 : 0.51 (\textbf{11\%}) & 0.48 : 0.40 (\textbf{17\%}) & 0.72 : 0.66 (\textbf{8\%}) & 0.65 : 0.56 (\textbf{14\%}) \\ \hline
\end{tabular}
\caption{Quantitative results of the SRH algorithm on the benchmark dataset \protect\cite{GONG14}. The quality of the results is evaluated w.r.t four attributes: texture, brokenness, colorfulness and softness. $E_r$ is the error for shadow region only, $E_r*$ is the error for the entire image. Results format~- Error before harmonization : Error after harmonization with SRH (relative error decrease).
}
\label{tab:perform}
\end{table*}

\subsection{Robustness and Parameter Settings} \label{sec:cp}

\textbf{Sensitivity to the initial result $I^N$} -
To evaluate the robustness of the SRH method, we manually generate synthetic examples of the initial result, $I^N$, with varying amount of residual shadow,
by linearly interpolating shadow images with ground-truth shadow-free images using a fixed alpha matte $\alpha$.
\figref{qr} illustrates the performance of the SRH method on one such example on images generated with different $\alpha$ values and reports the errors of the corrected results against the ground-truth.
For low values of $\alpha$ ($0.2$), i.e., medium shadow residuals, our algorithm can still generate a visually compelling result.
At high values of $\alpha$ ($0.5$), the initial result is significantly corrupted, and as expected, visual artifacts arise and errors increase.
We also evaluate the case of medium residual shadows ($\alpha = 0.2$) and a strong Gaussian noise.
Our algorithm successfully corrects the color distribution and texture details of the shadow region (\figref{qr}(g)), suggesting it is robust against noise due to the use of patch statistics.

\textbf{Color space and gain/bias settings} -
As described in \secref{model}, our shadow harmonization model uses CIELab color space, and enables gain in the $L$ channel and bias in the $a$ and $b$ channels for color correction. We denote this model as Model 0).
Here we compare its performance with other commonly-used color spaces and gain/bias settings:
Model 1 (CIELAB, enabling $L$ gain and $L$, $a$, $b$ bias);
Model 2 (RGB, enabling $R$, $G$, $B$ gain);
Model 3 (RGB, enabling $R$, $G$, $B$ gain and bias);
Model 4 (HLS, enabling $L$ gain and $H$, $S$ bias).
For each model, \equref{con1} and \equref{con2} are modified depending on whether gain/bias is enabled for each channel.
If both gain and bias are enabled for a channel, the two values are computed by matching the mean and standard variation of the patch pairs.
We compare these different color models on the benchmark dataset, using \cite{GONG14}'s shadow removal results as $I^N$.
Resulting errors of each model are reported in supplemental material, which suggest that Model 0 achieves slightly better performance than other models.


\textbf{Parameter ranges}
Parameter range settings (gain of L channel, bias of a, b channel) are important parameters in the patch-based synthesis process of SRH (\secref{pbs}).
If the ranges are too narrow, PatchMatch may not have sufficient freedom to correct errors in the initial shadow removal results. On the other hand, if the ranges are too wide, patches of different materials are more likely to be matched.
To find the best parameter ranges, we test different range settings on the benchmark dataset, again using \cite{GONG14}'s shadow removal results as $I^N$.
Specifically, we test three parameter range settings: narrow ($L$ gain $[0.99, 1.01]$, $a$ \& $b$ bias $[-0.01, 0.01]$), middle ($L$ gain $[0.9, 1.11]$, $A$ \& $b$ bias $[-0.05, 0.05]$), and wide ($L$ gain $[0.8, 1.25]$, $a$ \& $b$ bias $[-0.1, 0.1]$).
The resulting errors are reported in supplemental material in detail.
The middle range setting achieves the best performance and we fix the range to this range when generating all results reported in the paper. Also, the performance difference between the three settings are small, indicating the robustness of the algorithm against these parameter settings.

\subsection{Limitations} \label{sec:lim}

\begin{figure}[tb]
\setlength{\tabcolsep}{1pt}
    \centering
\begin{tabular}{c@{\hspace{0.05in}}c@{\hspace{0.05in}}c}
\includegraphics[width = 0.3\columnwidth]{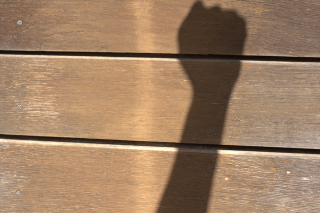} &
\includegraphics[width = 0.3\columnwidth]{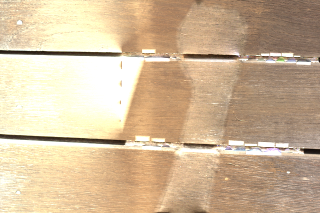} &
\includegraphics[width = 0.3\columnwidth]{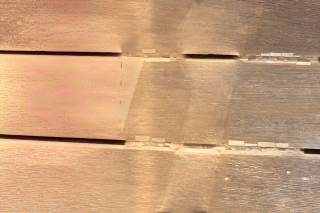} \\
\includegraphics[width = 0.3\columnwidth]{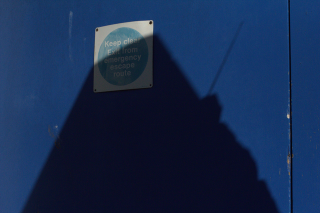} &
\includegraphics[width = 0.3\columnwidth]{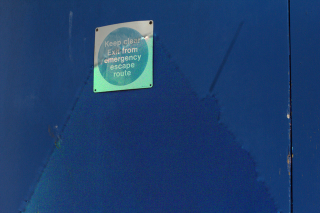} &
\includegraphics[width = 0.3\columnwidth]{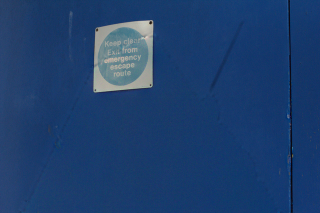} \\
(a) Input image & (b) Initial result 
& (c) Our result
\end{tabular}
    \caption{Failure cases. (top) Our algorithm does not handle significant shadow detection failure as shown in this example from \protect\cite{ICME2013}. (bottom) The SRH algorithm harmonizes the interior of the shadow region, but cannot fix the significant boundary discontinuities on this example from \protect\cite{GONG14}.
    }
    \label{fig:fc}
\end{figure}

Our approach has several limitations. Firstly, the SRH method cannot correct errors introduced by shadow detection failure.
For the example showed in \figref{fc}(top), the shadow detection process of \cite{ICME2013} fails and generates removal results with strong artifacts in the non-shadow region.
SRH removes some of them, but noticeable artifacts still persist.

We have also observed that previous shadow removal methods sometimes generate significant boundary discontinuities. \figref{fc}(bottom) shows one such example. Our method is mainly designed for harmonizing the interior of the shadow region, and is more limited at correcting such boundary discontinuities. As shown in the figure, it successfully removes color and texture inconsistencies inside the shadow region, but leaves some amount of boundary discontinuity in the final result.


\section{Conclusion}

We propose a fully automatic Shadow Region Harmonization approach for removing color and texture inconsistencies introduced by previous shadow removal methods. This technique is based on a parametric correction model, whose parameters are estimated by reconstructing the shadow region using non-shadow patches through a patch-based, guided image synthesis process.
We also introduce synthesis confidence to deal with unique structures and materials inside the shadow region.
Extensive evaluation shows the effectiveness and robustness of the proposed method.
{\small
\bibliographystyle{ieee}
\bibliography{template}

\begin{thebibliography}{10}\itemsep=-1pt

\bibitem{Baba:2004}
M.~Baba, M.~Mukunoki, and N.~Asada.
\newblock Shadow removal from a real image based on shadow density.
\newblock In {\em ACM SIGGRAPH 2004 Posters}, 2004.

\bibitem{Barnes:2009:PatchMatch}
C.~Barnes, E.~Shechtman, A.~Finkelstein, and D.~B. Goldman.
\newblock Patchmatch: a randomized correspondence algorithm for structural
  image editing.
\newblock {\em ACM Trans. Graph.}, 28(3), 2009.

\bibitem{Barnes:2010:GPatchMatch}
C.~Barnes, E.~Shechtman, D.~B. Goldman, and A.~Finkelstein.
\newblock The generalized patchmatch correspondence algorithm.
\newblock In {\em Proc. ECCV}, 2010.

\bibitem{RIS78}
H.~G. Barrow and J.~M. Tenenbaum.
\newblock Recovering intrinsic scene characteristics from images.
\newblock In {\em Int. Conf. on Computer Vision Systems}, 1978.

\bibitem{Darabi12:ImageMelding12}
S.~Darabi, E.~Shechtman, C.~Barnes, D.~B. Goldman, and P.~Sen.
\newblock {I}mage {M}elding: Combining inconsistent images using patch-based
  synthesis.
\newblock {\em ACM Trans. Graph.}, 31(4), 2012.

\bibitem{SFS14}
A.~Ecins, C.~Fermüller, and Y.~Aloimonos.
\newblock {\em Shadow free segmentation in still images using local density
  measure}.
\newblock IEEE Computer Society, 2014.

\bibitem{eccv:2004}
G.~Finlayson, M.~Drew, and C.~Lu.
\newblock Intrinsic images by entropy minimization.
\newblock In {\em Proc. ECCV}, 2004.

\bibitem{02ECCV}
G.~D. Finlayson, S.~D. Hordley, C.~Lu, and M.~S. Drew.
\newblock Removing shadows from images.
\newblock In {\em Proc. ECCV}, 2002.

\bibitem{05Hami}
C.~Fredembach and G.~D. Finlayson.
\newblock Hamiltonian path-based shadow removal.
\newblock In {\em Proc. BMVC}, 2005.

\bibitem{GONG14}
H.~Gong and D.~Cosker.
\newblock Interactive shadow removal and ground truth for variable scene
  categories.
\newblock In {\em Proc. BMVC}. BMVA Press, 2014.

\bibitem{ICME2013}
H.~Gong, D.~Cosker, C.~Li, and M.~Brown.
\newblock User-aided single image shadow removal.
\newblock In {\em Proc. ICME}, 2013.

\bibitem{grosse09intrinsic}
R.~Grosse, M.~K. Johnson, E.~H. Adelson, and W.~T. Freeman.
\newblock Ground-truth dataset and baseline evaluations for intrinsic image
  algorithms.
\newblock In {\em Proc. ICCV}, 2009.

\bibitem{Gryka2015softShadows}
M.~Gryka, M.~Terry, and G.~J. Brostow.
\newblock Learning to remove soft shadows.
\newblock {\em ACM Trans. Graph.}, 2015.

\bibitem{11cvpr}
R.~Guo, Q.~Dai, and D.~Hoiem.
\newblock Single-image shadow detection and removal using paired regions.
\newblock In {\em Proc. CVPR}, 2011.

\bibitem{HaCohen:2011:NRDC}
Y.~HaCohen, E.~Shechtman, D.~B. Goldman, and D.~Lischinski.
\newblock Non-rigid dense correspondence with applications for image
  enhancement.
\newblock {\em ACM Trans. Graph.}, 30(4), 2011.

\bibitem{Hertzmann:2001}
A.~Hertzmann, C.~E. Jacobs, N.~Oliver, B.~Curless, and D.~H. Salesin.
\newblock Image analogies.
\newblock In {\em Proc. ACM SIGGRAPH}, 2001.

\bibitem{eccv:2008}
F.~Liu and M.~Gleicher.
\newblock Texture-consistent shadow removal.
\newblock In {\em Proc. ECCV}, 2008.

\bibitem{ISR:09}
D.~Miyazaki, Y.~Matsushita, and K.~Ikeuchi.
\newblock Interactive shadow removal from a single image using hierarchical
  graph cut.
\newblock In {\em Proc. ACCV}, volume 5994. 2010.

\bibitem{ESS:07}
A.~Mohan, J.~Tumblin, and P.~Choudhury.
\newblock Editing soft shadows in a digital photograph.
\newblock {\em IEEE Computer Graphics and Applications}, 27(2), 2007.

\bibitem{Sanin:2012}
A.~Sanin, C.~Sanderson, and B.~C. Lovell.
\newblock Shadow detection: A survey and comparative evaluation of recent
  methods.
\newblock {\em Pattern Recogn.}, 45(4), 2012.

\bibitem{CGF08}
Y.~Shor and D.~Lischinski.
\newblock The shadow meets the mask: Pyramid-based shadow removal.
\newblock {\em Computer Graphics Forum}, 27(2), 2008.

\bibitem{Wexler:2007:SCV}
Y.~Wexler, E.~Shechtman, and M.~Irani.
\newblock Space-time completion of video.
\newblock {\em IEEE Trans. PAMI}, 29(3), 2007.

\bibitem{05Bys}
T.-P. Wu and C.-K. Tang.
\newblock A bayesian approach for shadow extraction from a single image.
\newblock In {\em Proc. ICCV}, volume~1, 2005.

\bibitem{tog07}
T.-P. Wu, C.-K. Tang, M.~S. Brown, and H.-Y. Shum.
\newblock Natural shadow matting.
\newblock {\em ACM Trans. Graph.}, 26(2), 2007.

\bibitem{Xiao132}
C.~Xiao, D.~Xiao, L.~Zhang, and L.~Chen.
\newblock Efficient shadow removal using subregion matching illumination
  transfer.
\newblock {\em Computer Graphics Forum}, 32(7), 2013.

\bibitem{Xu06shadowdetection}
L.~Xu, F.~Qi, R.~Jiang, Y.~Hao, G.~Wu, L.~Xu, F.~Qi, R.~Jiang, Y.~Hao, and
  G.~Wu.
\newblock Shadow detection and removal in real images: A survey, 2006.

\end{thebibliography}
}

\end{document}